\documentclass[11pt]{article}

\usepackage[style=apa, backend=biber]{biblatex}
\usepackage[letterpaper,margin=1in]{geometry}
\usepackage{amsmath,graphicx,hyperref,titling}
\hypersetup{colorlinks=true,allcolors=blue}

\pretitle{\centering\LARGE\bfseries}
\posttitle{\par\vspace{1em}}

\preauthor{\centering\large}
\postauthor{\par\vspace{1em}}

\predate{\centering\normalsize}
\postdate{\par\vspace{2em}}

\title{Self-Consistency Is Losing Its Edge: Diminishing Returns and Rising Costs in Modern LLMs}
\author{Chiyan Loo \\ \texttt{loochiyan@gmail.com}}
\date{October 28, 2025}

\begin{document}
\maketitle

\renewcommand{\abstractname}{\large\bfseries Abstract}

\begin{abstract}
Self-consistency \parencite{wang2022self}---sampling multiple reasoning paths and selecting the most frequent answer---was designed for an era when language models made frequent, unpredictable errors. This study argues that the technique has become increasingly wasteful as models grow stronger, and may degrade performance on problems that modern models already solve reliably. Using Gemini 2.5 models on HotpotQA \parencite{yang2018hotpotqa} and MATH-500 \parencite{hendrycks2021measuring}, we show that accuracy gains from increasing the number of sampled reasoning paths are minimal---0.4\% on HotpotQA across 20 samples, and 1.6\% on MATH-500---while token costs scale nearly linearly with sample count. Critically, performance plateaued early and in some configurations \emph{declined} at high sample counts, suggesting that additional paths introduce noise rather than signal when models already solve problems reliably. As inference costs rise with model scale, indiscriminate self-consistency is difficult to justify. We recommend reserving multi-path sampling for problems that demonstrably exceed a model's single-pass reliability.
\end{abstract}

\section{Introduction}

Self-consistency \parencite{wang2022self} improves reasoning reliability in large language models (LLMs) by sampling multiple reasoning paths and selecting the most consistent answer. Originally proposed for weaker models, it mitigated stochastic reasoning errors through aggregation. Multi-agent reasoning generalizes this idea by allowing several independent reasoning processes to generate and compare trajectories, seeking higher accuracy and interpretability.

The core justification for self-consistency was high model variance: when individual outputs frequently disagree, aggregation extracts a more reliable signal. That justification has weakened considerably. Frontier models have become dramatically more capable, and problems that once challenged them---multi-hop factual retrieval, arithmetic, symbolic reasoning---are now solved reliably in a single pass. When a model already answers correctly most of the time, additional sampled paths are near-identical and add no new information. Worse, they introduce occasional spurious errors that the aggregator cannot always filter out.

These costs are not abstract. Modern frontier models charge substantially more per token than their predecessors, and configurations with many sampled paths multiply token usage linearly. This study revisits self-consistency using modern LLMs to evaluate whether drawing more reasoning samples still yields meaningful benefits, or whether the technique has become an expensive habit mismatched to current model capabilities.

Results confirm that accuracy gains plateau early and, in some configurations, \emph{decline} at high sample counts---a pattern inconsistent with diminishing returns alone and more consistent with noise introduction on problems that were already solved. This suggests self-consistency should be reserved for genuinely difficult problems rather than applied as a default scaling strategy.

\section{Related Work}

Early work on reasoning with large language models demonstrated that combining chain-of-thought (CoT) prompting \parencite{wei2022chain} with majority-vote aggregation over multiple sampled paths improved accuracy significantly \parencite{wang2022self}. This established the efficacy of multi-path reasoning when individual model outputs were unreliable.

Subsequent research recognized the efficiency problems with simple self-consistency. \textcite{aggarwal2023let} introduced Adaptive-Consistency, dynamically halting sampling once answers converge, reducing sample usage by up to 7.9$\times$ while dropping accuracy by less than 0.1\%. \textcite{wan2024reasoning} proposed criteria-based early stopping, cutting sample usage by roughly 70\% with similarly negligible accuracy loss. Both works implicitly acknowledge that there is a sampling threshold beyond which additional paths are redundant.

What neither work fully addresses is that this threshold shifts dramatically as model capability improves. For a model that already solves 98\% of MATH-500 problems in a single pass, the threshold is effectively one sample for most problems. The present work makes this explicit: as models grow stronger and tasks become easier, self-consistency transitions from a useful correction mechanism to an expensive source of noise.

\section{Methodology}

\subsection{Experimental Setup}
This study adopted a structured self-consistency framework to evaluate the marginal benefit of increasing the number of sampled reasoning paths in modern LLMs. Each configuration drew multiple independent CoT \parencite{wei2022chain} responses from the same model. A secondary aggregator model then identified the most internally consistent answer among the resulting traces.

We tested sample counts of 3, 5, 10, 15, and 20 reasoning paths. A single-sample CoT baseline served as the control to isolate the contribution of multi-path sampling. Temperature, top-$p$, and maximum tokens were held constant across configurations. Prompts and evaluation contexts were identical across runs to avoid ordering bias, and the aggregation step was deterministic.

\subsection{Datasets}
Two benchmarks were selected to represent distinct reasoning domains:

\begin{itemize}
  \item \textbf{HotpotQA} \parencite{yang2018hotpotqa}: A multi-hop question answering dataset requiring integration of evidence across documents, testing logical composition and factual consistency.
  \item \textbf{MATH-500} \parencite{hendrycks2021measuring}: Mathematics problems spanning arithmetic, algebra, geometry, and symbolic reasoning, assessing step-by-step deductive reasoning.
\end{itemize}

Both datasets historically motivated self-consistency research but have become substantially easier for frontier models, making them well-suited to reveal ceiling effects and redundancy dynamics.

\subsection{Evaluation Metrics}
\begin{description}
  \item[Accuracy:] Outputs were compared against reference answers using an evaluator LLM scoring on semantic equivalence rather than surface form.
  \item[Cost:] Total token consumption was recorded for each configuration, including all sampled outputs and aggregator reasoning, to capture the compute overhead of self-consistency at each sampling scale.
\end{description}

\subsection{Procedure}
Each sample was processed under all experimental configurations. For the baseline, a single CoT output was generated and evaluated directly. For multi-path conditions, independent traces were generated in parallel and reviewed by the aggregator. Accuracy and token cost were aggregated and visualized as accuracy--cost tradeoff curves.

\section{Results}

Across both datasets, accuracy gains from self-consistency were small and plateaued early, while costs scaled linearly. In some configurations performance actually declined at high sample counts---a more troubling finding than simple diminishing returns.

\begin{figure}[ht]
\centering
\includegraphics[width=1\textwidth]{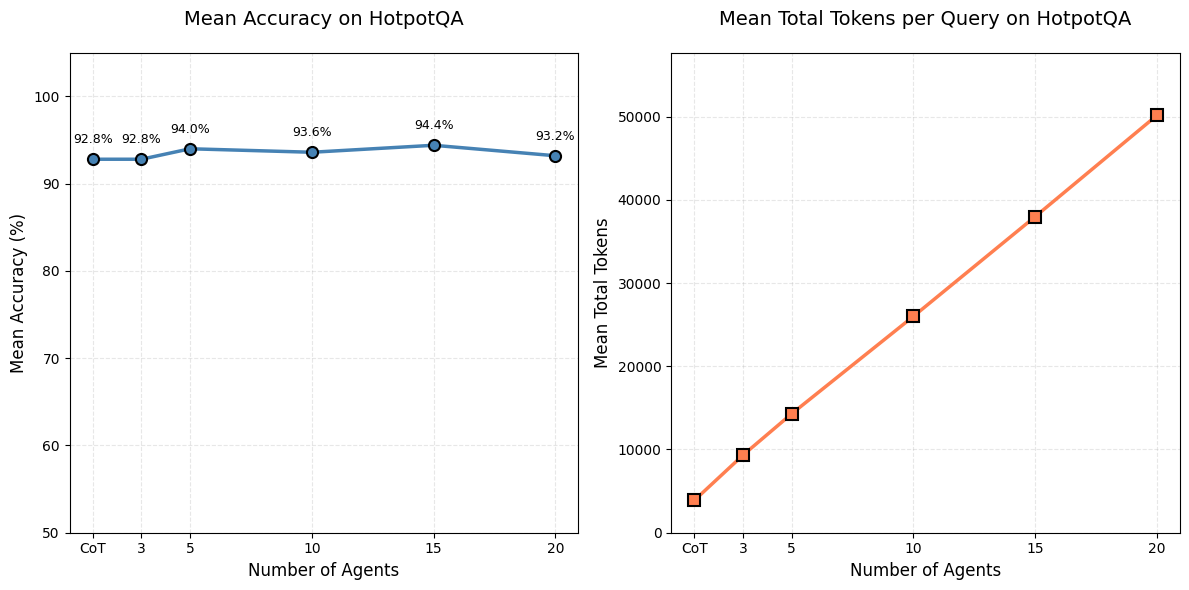}
\caption{Gemini-2.5-Flash-Lite accuracy and cost on HotpotQA. The 0.4\% total gain across 20 sampled reasoning paths does not justify the approximately 20$\times$ increase in token cost.}
\label{fig:hotpotqa}
\end{figure}

On HotpotQA \parencite{yang2018hotpotqa}, Gemini-2.5-Flash-Lite improved by only 0.4\% between the single CoT baseline and the 20-sample configuration. As shown in Figure~\ref{fig:hotpotqa}, accuracy fluctuated irregularly rather than rising steadily---the signature of a model already near ceiling where additional sampled paths vary around a high baseline rather than correcting systematic errors. Token usage, meanwhile, scaled nearly linearly, making the implied cost per accuracy point extremely high.

\begin{figure}[ht]
\centering
\includegraphics[width=1\textwidth]{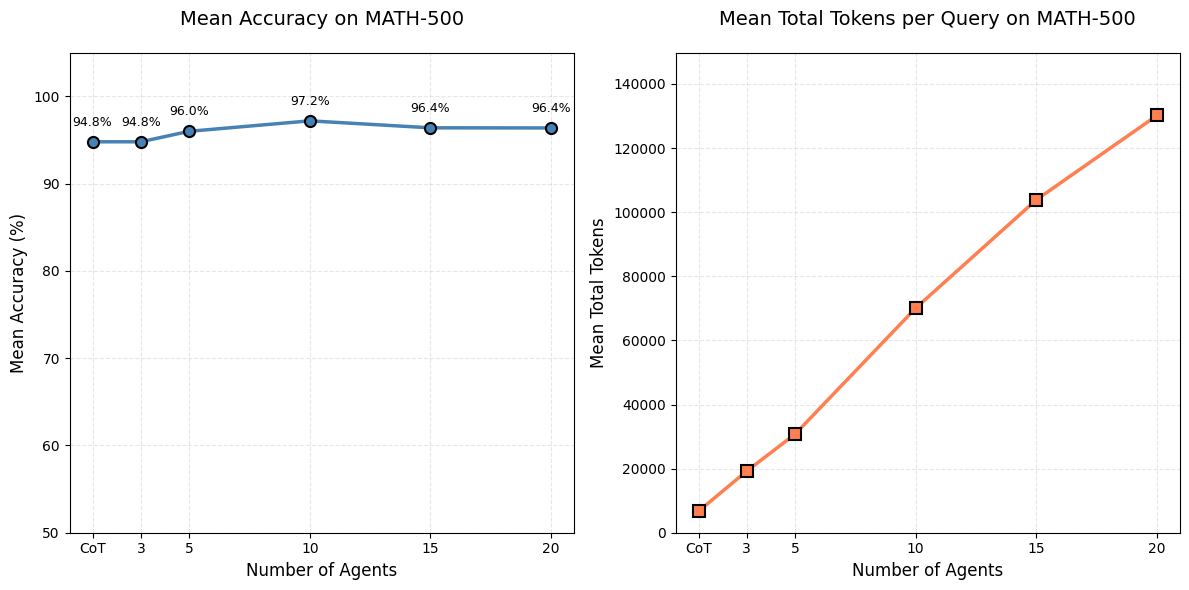}
\caption{Gemini-2.5-Flash-Lite accuracy and cost on MATH-500. Accuracy peaks around 10 sampled paths and declines beyond 15, while cost continues scaling linearly.}
\label{fig:math-500}
\end{figure}

On MATH-500 \parencite{hendrycks2021measuring}, Flash-Lite accuracy improved through approximately 10 sampled paths before plateauing and then \emph{declining} slightly beyond 15, as shown in Figure~\ref{fig:math-500}. This decline is notable: it suggests that once a model reliably solves most problems, additional samples introduce occasional wrong reasoning paths that the aggregator cannot fully suppress. Problems that were easy become harder to handle correctly in aggregate when a few incorrect paths pollute the pool.

\begin{figure}[ht]
\centering
\includegraphics[width=1\textwidth]{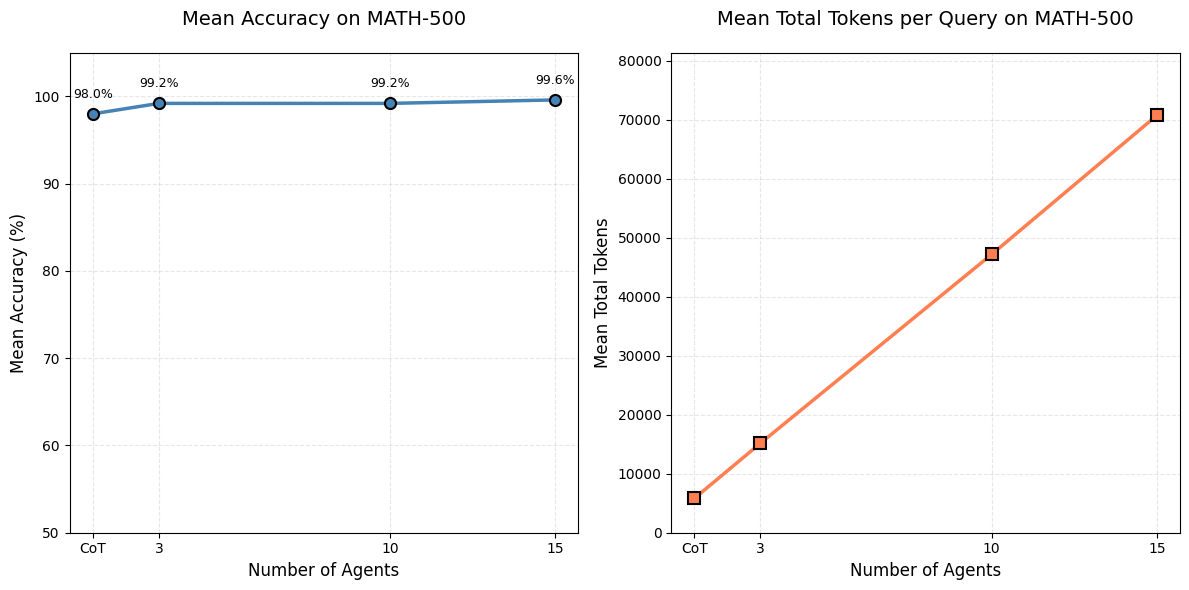}
\caption{Gemini-2.5-Pro accuracy and cost on MATH-500. Starting from 98\%, the model gains 1.6\% over 15 sampled reasoning paths at roughly 15$\times$ the compute cost.}
\label{fig:pro}
\end{figure}

Gemini-2.5-Pro, tested on MATH-500 with up to 15 sampled reasoning paths, began at a 98\% CoT baseline and improved to 99.2\% at 3 paths and 99.6\% at 15---a total gain of 1.6\% at approximately 15$\times$ the single-sample token cost. As shown in Figure~\ref{fig:pro}, its curve was smoother than Flash-Lite's, reflecting stronger internal coherence. But this coherence is precisely the problem: a model whose outputs already agree closely has little variance for aggregation to exploit. The smoother curve signals that self-consistency had little work to do, not that it was more effective.

Overall, both models confirmed that self-consistency yields rapidly diminishing---and occasionally negative---returns as the number of sampled paths grows, while compute costs increase without bound.

\section{Discussion and Conclusion}

Past work \parencite{wang2022self} demonstrated strong accuracy gains from self-consistency when models exhibited high output variance and errors were frequent. Our experiments with Gemini 2.5 models reproduce the plateau pattern but at a much smaller scale of improvement, and with a more troubling finding: accuracy can \emph{decline} at high sample counts, as seen with Flash-Lite on MATH-500 beyond 15 sampled paths. This is not mere diminishing returns---it is evidence that self-consistency can actively degrade performance when models are too capable for the task.

The underlying cause is redundancy. When a model already solves most problems correctly in a single pass, additional sampled paths are near-identical. Aggregating them does not improve accuracy; it introduces the occasional spurious error that the aggregator fails to suppress. Gemini-2.5-Pro's 98\% CoT baseline on MATH-500 illustrates the extreme case: 15 sampled paths cost 15$\times$ the compute to recover 1.6\% on problems almost entirely already solved. The vast majority of that spend confirms answers the model already had correct.

As frontier model inference costs continue to rise, this inefficiency carries real consequences. Drawing 20 reasoning samples costs roughly 20$\times$ the tokens of a single CoT pass---a multiplier that is difficult to justify for gains under 2\% on problems modern models largely solve unaided. Self-consistency remains a valid technique, but it belongs in a narrow role: reserved for problems that demonstrably exceed a model's single-pass reliability, not applied as a default scaling strategy. Practitioners should evaluate baseline single-pass accuracy before committing to high sample counts, and treat self-consistency as a targeted tool rather than an automatic one.

\section{Limitations and Future Work}

This study's conclusions are constrained by its limited scope. Only 250 rows were evaluated for Gemini-2.5-Flash-Lite with up to 20 sampled reasoning paths, and Gemini-2.5-Pro was tested solely on MATH-500 with a maximum of 15. These restrictions limit generalizability across tasks, model families, and problem difficulties.

Critically, both benchmarks may now be too easy for the evaluated models, which is precisely what makes them useful for demonstrating ceiling effects but limits insight into where self-consistency remains beneficial. Future work should identify problem categories that reliably sit in the range where single-pass accuracy is meaningfully below ceiling---where model outputs exhibit genuine variance---and quantify the sampling count at which multi-path aggregation becomes worthwhile. Developing difficulty-aware routing that applies self-consistency selectively only to hard queries would be a practically valuable contribution.

\section*{Acknowledgments}
The author utilized AI tools for assistance in drafting, structuring, and technical editing of this manuscript. The content was reviewed and the core experimental design, data analysis, citations, and conclusions were developed by the author.

\newpage
\printbibliography

\end{document}